\documentclass[sigconf]{acmart}
\usepackage{comment}
\usepackage{algorithm}
\usepackage{algorithmic}
\usepackage{bm}
\usepackage{amsthm}
\usepackage{amsmath}
\usepackage{nicefrac}
\usepackage{soul}
\usepackage{pifont}
\usepackage{subfigure}
\usepackage{multirow}
\usepackage[thinc]{esdiff}
\usepackage{balance}

\usepackage[normalem]{ulem}

\usepackage{pdftexcmds}
\usepackage{catchfile}
\usepackage{ifluatex}
\usepackage{ifplatform}
\usepackage{hyperref}
\usepackage{enumitem}

\newtheorem{remark}{Remark}

\newcommand{\model}{DiffMG}

\theoremstyle{plain}
\newtheorem{proposition}{Proposition}

\theoremstyle{definition}
\newtheorem{definition}{Definition}[section]

\AtBeginDocument{%
  \providecommand\BibTeX{{%
    \normalfont B\kern-0.5em{\scshape i\kern-0.25em b}\kern-0.8em\TeX}}}

\copyrightyear{2021} 
\acmYear{2021} 
\setcopyright{acmcopyright}
\acmConference[KDD '21]{Proceedings of the 27th ACM SIGKDD Conference on Knowledge Discovery and Data Mining}{August 14--18, 2021}{Virtual Event, Singapore}
\acmBooktitle{Proceedings of the 27th ACM SIGKDD Conference on Knowledge Discovery and Data Mining (KDD '21), August 14--18, 2021, Virtual Event, Singapore}
\acmPrice{15.00}
\acmDOI{10.1145/3447548.3467447}
\acmISBN{978-1-4503-8332-5/21/08}

\acmSubmissionID{rst3415}

\begin{document}
\fancyhead{}

\title{DiffMG: Differentiable Meta Graph Search for Heterogeneous Graph Neural Networks}

\author{
    Yuhui Ding\textsuperscript{\rm 1}, 
    Quanming Yao\textsuperscript{\rm 2,3}, 
    Huan Zhao\textsuperscript{\rm 2},
    Tong Zhang\textsuperscript{\rm 1}
}

\affiliation{
    \textsuperscript{\rm 1}Department of Computer Science and Engineering, Hong Kong University of Science and Technology \\
    \textsuperscript{\rm 2}4Paradigm Inc. 
    \textsuperscript{\rm 3}Department of Electronic Engineering, Tsinghua University \\
    yuhui.ding@connect.ust.hk,
    \{yaoquanming, zhaohuan\}@4paradigm.com,
    tongzhang@ust.hk
    \country{}
}

\begin{abstract}
In this paper, we propose a novel framework to automatically utilize task-dependent semantic information which is encoded in heterogeneous information networks (HINs). Specifically,
we search for a meta graph, which can capture more complex semantic relations than a meta path, to determine how graph neural networks (GNNs) propagate messages along different types of edges. We formalize the problem within the framework of neural architecture search (NAS) and then perform the search in a differentiable manner. We design an expressive search space in the form of a directed acyclic graph (DAG) to represent candidate meta graphs for a HIN, 
and we propose task-dependent type constraint to filter out those edge types along which message passing has no effect on the representations of nodes that are related to the downstream task. The size of the search space we define is huge,
so we further propose a novel and efficient search algorithm to make the total search cost on a par with training a single GNN once. Compared with existing popular NAS algorithms,
our proposed search algorithm improves the search efficiency. We conduct extensive experiments on different HINs and downstream tasks to evaluate our method, and experimental results show that our method can outperform state-of-the-art heterogeneous GNNs and also improves efficiency compared with those methods which can implicitly learn meta paths\footnote{
Code is available at: \url{https://github.com/AutoML-4Paradigm/DiffMG}.
Q. Yao is the corresponding author.
}.
\end{abstract}



\maketitle

\section{Introduction}
Heterogeneous information networks (HINs) are ubiquitous in real-world applications, 
such as web mining~\cite{zhang2019oag}, recommender systems~\cite{jin2020efficient}
and biomedicine~\cite{zitnik2019machine},
to model complex relations among individual entities.
Alongside the network topology,
a HIN also encodes rich semantic information
which consists of multiple node types, edge types
and node features which come from 
different sources.
For example,
Microsoft Academic Graph~\cite{wang2020microsoft} has multiple 
types of nodes: authors, papers, venues, etc.,
and multiple types of relations among them
(see Figure~\ref{fig:schema}).
Such additional semantic information increases the difficulty of representation learning
on HINs, compared with homogeneous networks~\cite{wang2019heterogeneous,NIPS2019_9367,fu2020magnn, zhang2018deep}.

Recently,
graph neural networks (GNNs),
which typically fall into a common message passing framework \cite{gilmer2017neural},
have achieved remarkable progress 
in many graph-based machine learning tasks~\cite{kipf2016semi, gilmer2017neural, zhang2018link}.
Within this framework, each node generates its representation by aggregating features 
of its neighbors.
Despite the success of GNNs on homogeneous networks,
it has been observed that, naive message passing on HINs which ignores node types and edge types often leads to sub-optimal performance~\cite{wang2019heterogeneous,NIPS2019_9367}.
To utilize semantic information encoded in HINs,
MAGNN~\cite{fu2020magnn} and HAN~\cite{wang2019heterogeneous} are designed specifically for HINs.
They employ meta paths~\cite{sun2011pathsim} to define what neighbors to aggregate information from. 
However, it can be extremely difficult to design meta paths by hand 
with little or no domain knowledge~\cite{yang2018similarity}, especially for complicated heterogeneous systems (e.g., various types of 
chemical atoms).

Other heterogeneous GNN variants,
e.g., GTN~\cite{NIPS2019_9367} and HGT~\cite{hu2020heterogeneous},
implicitly learn meta paths by fusing
different types based on attention mechanisms.
However, such methods face two potential issues.
First, due to the heterogeneity,
not all semantic information is useful to the downstream task.
For example,
when we want to predict research areas for authors in an academic network
(Figure~\ref{fig:schema}),
institutions that authors are affiliated with are unrelated
information and may bring noise to author representations.
Second,
mechanisms used to fuse different node types and edge types increase model
complexity and harm the efficiency.
For example,
HGT~\cite{hu2020heterogeneous}
adapts the transformer architecture~\cite{NIPS2017_3f5ee243} and 
keeps distinct hidden weight matrices for each node type and each edge type
at each layer to compute attention scores,
which makes it much less efficient than homogeneous counterparts.

Moreover,
above methods only consider meta paths,
which have limited capacity to capture intricate semantic proximity
compared with more expressive meta graphs~\cite{huang2016meta, zhao2017meta}.
A more recent method GEMS~\cite{han2020genetic} utilizes 
the evolutionary algorithm to search for meta graphs between source node types
and target node types for recommendation,
but it is too slow to fully explore the search space within a reasonable 
time budget.
Therefore,
it has not been well addressed yet how
GNNs can automatically utilize task-dependent semantic information in HINs
and at the same time keep high efficiency (see Table~\ref{tab:hingnn} for a summary).

\begin{table}[t]
\small
	\caption{Comparison of heterogeneous GNNs.
		Semantics:
		what kind of semantic information a method utilizes.
		Construction: how a method finds the semantics.
		}
	\vspace{-10px}
	\label{tab:hingnn}
	\begin{tabular}{c | c | c }
		\toprule
		          Method                 & Semantics  & Construction  \\ \midrule
		HAN~\cite{wang2019heterogeneous} & Meta path  & Manual design  \\ \midrule
		    MAGNN~\cite{fu2020magnn}     & Meta path  & Manual design  \\ \midrule
		    GTN~\cite{NIPS2019_9367}     & Meta path  & Implicitly learn by attention      \\ \midrule
		 HGT~\cite{hu2020heterogeneous}  & Meta path  & Implicitly learn by attention      \\ \midrule
		   GEMS~\cite{han2020genetic}    & Meta graph & Genetic search         \\ \midrule
		       DiffMG                    & Meta graph & Differentiable search         \\ \bottomrule
	\end{tabular}
\end{table}

In this paper,
we propose a novel method to address the above issue,
namely \model~(\underline{Diff}erentiable \underline{M}eta \underline{G}raph search).
We are inspired by recent advancements in one-shot neural architecture search (NAS)~\cite{elsken2019neural, liu2018darts}
which automatically discovers novel promising CNN architectures.
Within the context of HIN,
\model~searches for a meta graph in a differentiable fashion
to guide how GNNs propagate messages along different types of edges.
\model~is trained in a two stage manner.
At the search stage,
\model~defines a search space
in the form of a directed acyclic graph (DAG) to represent a candidate meta graph,
and then architecture parameters are introduced to weight candidate edge types
for each link in the DAG search space.
Moreover,
we apply task-dependent type constraint to particular 
links in the DAG,
in order to filter out those candidate edge types along which message passing has no 
effect on the representations of nodes which are 
related to the downstream task.
The architecture parameters are then optimized end to end through bi-level optimization~\cite{franceschi2018bilevel}.
At the evaluation stage,
we derive a meta graph by selecting the most promising edge type for each link in the search space,
indicated by architecture parameters,
and then \model~is trained with the derived meta graph.
To improve the search efficiency,
we propose an efficient search algorithm,
so that only one edge type, instead of all candidates,
for each link in the DAG is involved in the computation 
per iteration during the search.
We evaluate \model~extensively on different heterogeneous datasets,
and experimental results show that \model~outperforms state-of-the-art heterogeneous GNNs
and also improves efficiency compared with methods that 
learn meta paths implicitly~\cite{NIPS2019_9367, hu2020heterogeneous}.
To summarize,
the main contributions of our work are:
\begin{itemize}[leftmargin=*]
    \item We address the problem of utilizing task-dependent semantic information in HINs
    by searching for a meta graph within the framework of NAS;
    \item We design an expressive DAG search space to represent candidate meta graphs and propose
    task-dependent type constraint to filter out unrelated edge types;
    \item We propose an efficient search algorithm to make the search cost on a par with training
    a single GNN once;
    \item We conduct extensive experiments to demonstrate the effectiveness and efficiency of
    our proposed method.
\end{itemize}

\section{Preliminaries and Notations}
\label{sec:pre:hin}

\begin{definition}[Heterogeneous Information Network (HIN)~\cite{yang2020heterogeneous}]
A heterogeneous information network is a directed graph which has
the form $\mathcal{G} = \{\mathcal{V}, \mathcal{E}, \mathcal{T}, 
\mathcal{R}, 
f_{\mathcal{T}}, 
f_{\mathcal{R}}\}$,
where $\mathcal{V}$ denotes the set of nodes and $\mathcal{E}$ denotes 
the set of edges.
$f_{\mathcal{T}}: \mathcal{V} \rightarrow \mathcal{T}$ 
is a mapping function which maps each node $v \in \mathcal{V}$ 
to a node type $f_{\mathcal{T}}(v) \in \mathcal{T}$.
Similarly,
$f_{\mathcal{R}}$ maps each edge $e \in \mathcal{E}$ to an edge type
$f_{\mathcal{R}}(e) \in \mathcal{R}$.
We call $\mathcal{S} = \{\mathcal{T}, \mathcal{R}\}$ the network schema of 
a heterogeneous information network, with
$|\mathcal{T}| + |\mathcal{R}| > 2$.
\end{definition}

We use $\bm{A}_{r}$ to denote the adjacency matrix formed by the edges of type $r \in \mathcal{R}$,
and $\bm{\mathcal{A}}$ to denote the collection 
of all $\bm{A}_{r}$'s.
An edge type is determined by its source node type and target node type.
For example, in Figure~\ref{fig:schema}, $\bm{A}_{\textsf{PA}}$ represents 
directed edges from author nodes (\textsf{A}) to paper nodes (\textsf{P}), 
while $\bm{A}_{\textsf{AP}}$ represents directed edges from \textsf{P} to \textsf{A}. 
If the source node type and the target node type of an edge type $r$ are the same,
then $\bm{A}_{r}$ is symmetric. 

\begin{definition}[Meta Path~\cite{sun2011pathsim}]
\label{def:meta_path}
A meta path $P$ is a composite relation defined on $\mathcal{S}$ 
which consists of multiple edge types, i.e.,
$P = t_1 \xrightarrow{r_1} t_2 \xrightarrow{r_2} \dots \xrightarrow{r_l} t_{l+1}$,
where $t_1, \dots, t_{l+1} \in \mathcal{T}$ and $r_1, \dots, r_l \in \mathcal{R}$.
One meta path corresponds to many meta path instances in the underlying HIN.
\end{definition}

Definition~\ref{def:meta_graph} below offers a natural generalization of meta paths,
which allows the in-degree of 
each node type (except the source node type) to be larger than $1$.
Intuitively, meta graphs can represent more intricate semantic relations 
than meta paths.
For example, to determine whether two people are classmates or alumni~\cite{zhang2020mg2vec},
school information alone is not enough.
Class information is also required.


\begin{definition}[Meta Graph~\cite{zhao2017meta}]
\label{def:meta_graph}
A meta graph is a directed acyclic graph (DAG) on $\mathcal{S}$, 
with a single source
node type (i.e., with in-degree $0$) and a single sink (target) node type (i.e., with out-degree $0$).
\end{definition}

We use $N$ to denote the total number of nodes in the HIN
and $d$ to denote the hidden dimension.
We use uppercase bold letters (e.g., $\bm{X}$) to denote matrices, 
and lowercase bold letters (e.g., $\bm{z}$) to denote column vectors.

\section{Related Work}

\subsection{Heterogeneous GNNs}
\label{sec:rel:hin}
Heterogeneous GNNs are designed for HINs
to utilize semantic information which is ignored by homogeneous ones.
One category of 
heterogeneous GNNs~\cite{zhang2018deep, wang2019heterogeneous, fu2020magnn, jin2020efficient} employs hand-designed meta paths (Definition~\ref{def:meta_path}) to define neighbors.
HAN~\cite{wang2019heterogeneous} extracts multiple homogeneous sub-networks based on
different meta paths and uses semantic-level attention to combine
node representations from different sub-networks.
MAGNN~\cite{fu2020magnn} utilizes RotatE~\cite{sun2018rotate}
to encode intermediate nodes along each meta path instance and 
combines multiple meta paths like HAN.
NIRec~\cite{jin2020efficient} proposes an interaction model 
which is based on meta path guided neighborhood for recommendation.
Compared with these methods,
\model~does not require prior knowledge of domain specific rules.

Another category of heterogeneous GNNs~\cite{zhang2019heterogeneous, NIPS2019_9367, hu2020heterogeneous}
designs mechanisms to fuse information from different types of nodes,
in order to eliminate the need for hand-designed meta paths.
GTN~\cite{NIPS2019_9367} combines adjacency matrices of different edge types using
attention scores and learns a new meta path based graph via matrix multiplication,
but it suffers from huge memory cost.
HetGNN~\cite{zhang2019heterogeneous} also mixes information of different node types
by attention.
However,
HetGNN is designed for a scenario where each node is associated with multi-modal
contents (e.g., text and image), which is different from our setting,
and it employs bi-directional LSTM to encode different types of contents.
HGT~\cite{hu2020heterogeneous}
keeps distinct hidden weight matrices 
for each type at each layer and computes mutual attention scores to weight messages 
coming along different types of relations.
Different from these methods,
\model~can select task-dependent semantic information rather than fuse all candidate
node types and edge types together.
In this way,
\model~filters out potential noise which is unrelated to the downstream task
and also improves model efficiency.

Moreover,
none of the above methods takes meta graphs~\cite{huang2016meta,fang2016semantic,zhao2017meta,yang2018meta} into account,
which show stronger capacity to express finer-grained semantics~\cite{zhang2020mg2vec}.
In~\cite{yang2018meta},
a meta graph is assessed by the eigenvalues of the adjacency matrix it defines,
but this method of assessment does not consider the properties of downstream tasks.
GEMS~\cite{han2020genetic} 
utilizes the evolutionary algorithm to search for meta graphs between source nodes
and target nodes for recommendation on HINs.
However, new candidate meta graphs which are generated by evolution 
have to be retrained with the GNN module from scratch for evaluation,
which makes it inefficient.
Compared with them,
\model~searches for a task-dependent meta graph in a differentiable way,
and the total search cost is on a par with
training a single GNN once.


\subsection{Neural Architecture Search (NAS)}
\label{sec:rel:nas}

Neural architecture search (NAS)~\cite{elsken2019neural}
has become a promising way to automatically discover novel 
architectures which outperform hand-designed ones.
Both the search space and the search algorithm are crucial to the 
effectiveness and efficiency of NAS. 
\begin{itemize}[leftmargin=*]
\item 
The search space is defined as the set of all possible architectures. 
It is task-specific and should be general enough to cover existing models.
At the same time,
it should be tractable to avoid high search cost.
For example,
by exploring the transferable blocks across different tasks,
the search space of CNNs has evolved from a macro architecture~\cite{zoph2017neural} to a micro cell structure~\cite{liu2018darts},
which significantly reduces the search cost on large image datasets~\cite{zoph2018learning}.

\item
The search algorithm aims to find the desired architecture in a given space.
Pioneer works~\cite{real2019regularized,zoph2018learning}
evaluate candidate architectures through stand alone training and require thousands of GPU days to obtain a good convolutional
neural network. 
Later,
one-shot NAS methods~\cite{xie2018snas, liu2018darts, pham2018efficient, yao2019differentiable} have improved the search efficiency by orders of magnitude via parameter sharing,
i.e.,
candidate child neural networks share parameters of a single
super neural network.
\end{itemize}

More recently,
some works have extended NAS
to homogeneous GNNs~\cite{ijcai2020-195, you2020design,lai2020policy,zhao2021search}.
GraphNAS~\cite{ijcai2020-195} utilizes reinforcement learning to select proper 
GNN components
(e.g., aggregation function, number of attention heads).
GraphGym~\cite{you2020design} proposes a GNN design space and a GNN task space to 
comprehensively evaluate model-task combinations.
Policy-GNN~\cite{lai2020policy} uses reinforcement learning to determine 
the number of aggregation layers for each node.
SANE~\cite{zhao2021search} follows DARTS~\cite{liu2018darts} to search for combinations of aggregation functions.

However, how to design a search space for heterogeneous GNNs
which is aware of semantic information in HINs is challenging,
and how to efficiently search the given space is also a non-trivial problem.
Both problems have not been addressed by above works.

\begin{figure*}[ht]
	\centering
	\subfigure[An example academic network with network schema.]
	{\includegraphics[width=.38\textwidth]{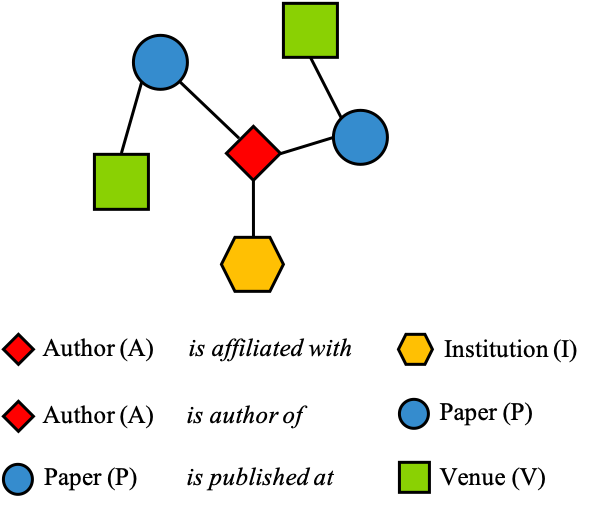}\label{fig:schema}}
	\quad
	\subfigure[Search space with $K = 2$.]
	{\includegraphics[width=.57\textwidth]{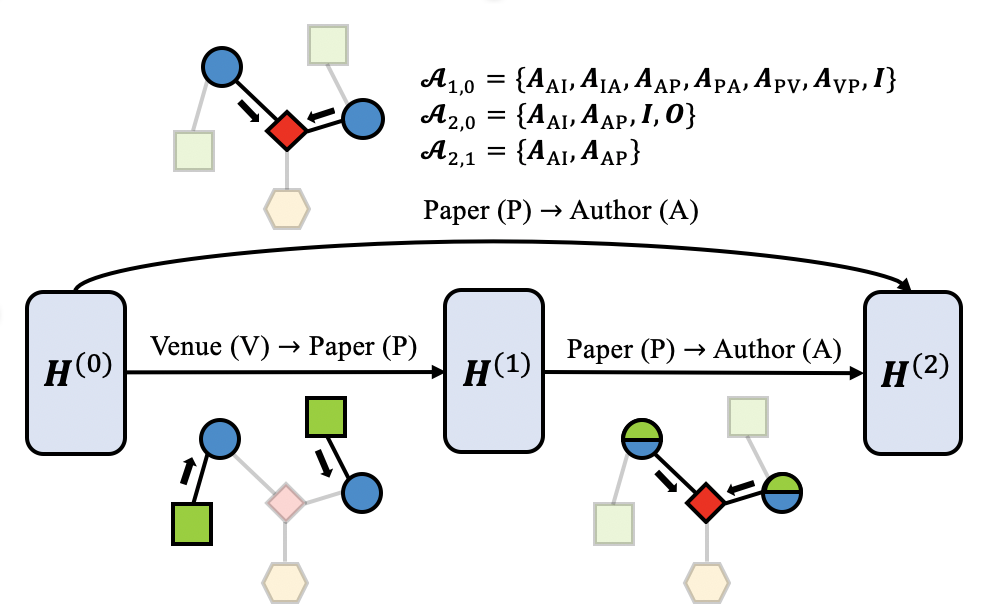}\label{fig:dag}}
	\vspace{-8px}
	\caption{
		Illustration of $F_{\bm{\mathcal{A}}}(\cdot)$ with $K = 2$ 
		for an example academic network (best viewed in color).
		Hidden weight matrices and non-linearity are omitted for ease of
		illustration.
		Here author (\textsf{A}) is the target node type related to
		the evaluation task which predicts research areas,
		and the derived meta graph does not propagate information 
		from \textsf{I} to \textsf{A}.
		Candidate edge types for each link in the search space are shown with task-dependent type constraint.
	}
	\label{fig:space}
\end{figure*}

\section{The Proposed Method}
\label{sec:method}

In this section,
we present the components of our proposed method in detail.
In Section~\ref{sec:space},
we define a search space to represent candidate meta graphs,
and describe how a meta graph in this search space determines message passing along different types of edges
in the underlying HIN.
Our search space has two key features:
\begin{itemize}[leftmargin=*]
\item First,
it is a DAG which allows more than one incoming link 
for each intermediate node (Section~\ref{sec:space:dag}).
This feature enhances its capacity to represent more complex semantics
than meta paths, i.e., meta graphs.

\item
Second,
it applies task-dependent type constraint
in order to filter out those edge types
along which message passing has no effect on the final representations of nodes which are related to the downstream task (Section~\ref{sec:space:A}).

\end{itemize}
Next,
in Section~\ref{sec:obj} we define the search objective, 
which presents a 
continuous relaxation of the designed search space.
Finally,
in Section~\ref{sec:alg} we
describe how \model~performs search in a 
differentiable way and propose an efficient
search algorithm.

In the sequel,
we will build our method based on graph convolutional network (GCN)~\cite{kipf2016semi},
which is the most popular GNN instance.
Other variants which also fall into the message passing framework,
like GraphSAGE~\cite{hamilton2017inductive}, GAT~\cite{velivckovic2017graph} and GIN~\cite{xu2018powerful}, 
can be employed and extended in a similar way.
We leave this for future work
and focus on searching for meta graphs in this paper.

\subsection{Search Space}
\label{sec:space}

Recall that 
a GCN layer 
for homogeneous networks
has the form:
\begin{equation}
\label{eq:gcn}
\bm{Z} = \sigma (\hat{\bm{A}}\bm{X}\bm{\Theta}),
\end{equation}
where $\bm{X} \in \mathbb{R}^{N \times d}$ and $\bm{Z} \in \mathbb{R}^{N \times d}$ denote the input
and output node representations of this layer, respectively. 
$\bm{\Theta}$ denotes a hidden weight matrix shared by all nodes, 
and $\sigma$ denotes the activation function.
$\hat{\bm{A}}$ denotes the normalized adjacency matrix which represents information propagation between 
homogeneous one-hop neighbors.
However,
for heterogeneous networks,
\eqref{eq:gcn} cannot differentiate different node types and thus loses semantic information.

To utilize the
semantic information,
it is important to differentiate messages which come along different edge types.
Moreover,
different combinations of messages from different edge types will generate 
different node representations.
Therefore,
we can construct a meta graph to describe the message passing process,
by selecting edge types and forming appropriate connections between them.
This motivates us
to extend~\eqref{eq:gcn} to the following heterogeneous message passing layer:
\begin{equation}
\label{eq:layer}
    \bm{Z} = \sigma \left( F_{\bm{\mathcal{A}}} \left( \bm{X}'\bm{\Theta} \right) \right ),
\end{equation}
where $\bm{X}'$ denotes the projected features output by type-specific transformation~\cite{wang2019heterogeneous, fu2020magnn} which 
projects features of different node types into a common latent space.
Here $F_{\bm{\mathcal{A}}}(\cdot)$ represents the message passing process 
which
is aware of the edge types,
i.e., $\bm{\mathcal{A}}$,
and can be described by a meta graph.

\subsubsection{Structure of $F_{\bm{\mathcal{A}}}(\cdot)$}
\label{sec:space:dag}
Specifically,
we define $F_{\bm{\mathcal{A}}}(\cdot)$ as a DAG,
in which 
ordered nodes denote intermediate representations in the message passing process.
The input of $F_{\bm{\mathcal{A}}}(\cdot)$ is denoted as $\bm{H}^{(0)}$ and the output
is $\bm{H}^{(K)}$, with $K$ the predefined number of intermediate states.
In this DAG, 
$\bm{H}^{(k)}$ ($1 \leq k \leq K$) is generated based on all its predecessors, i.e.,
\begin{equation}
\label{eq:H_k}
\bm{H}^{(k)} = \sum\nolimits_{0 \leq i < k}\Bar{f}_{k, i}(\bm{H}^{(i)}; \bm{\mathcal{A}}_{k, i}),
\end{equation}
where $\bar{f}_{k, i}(\cdot)$ denotes one message passing step which aggregates
$\bm{H}^{(i)}$ along edges of a certain type which is selected from $\bm{\mathcal{A}}_{k, i}$.
Here we choose graph convolution~\cite{kipf2016semi} as our aggregation function.

In Figure~\ref{fig:dag},
we illustrate a message passing process $F_{\bm{\mathcal{A}}}(\cdot)$ with $K = 2$
for an example academic network (Figure~\ref{fig:schema}).
$\bar{f}_{1,0}(\cdot)$ aggregates information from venue nodes to paper nodes,
and then $\bar{f}_{2,1}(\cdot)$ passes the aggregated information to the target author node.
At the same time,
$\bar{f}_{2,0}(\cdot)$ also passes information from paper nodes to the author node.
Note that both $\bar{f}_{2,1}(\cdot)$ and $\bar{f}_{2,0}(\cdot)$ choose the edge type
$\bm{A}_{\textsf{AP}}$ to propagate information, however,
the messages they pass are different because paper representations at the state $\bm{H}^{(1)}$
already aggregate information from venue nodes (mixing green and blue in Figure~\ref{fig:dag}).
This example shows that the DAG structure can flexibly combine multiple information propagation paths.

\subsubsection{Contents of $\bm{\mathcal{A}}_{k,i}$}
\label{sec:space:A}

Here we describe what is contained in each $\bm{\mathcal{A}}_{k,i}$.
In addition to edge types 
of the original HIN, i.e., $\bm{\mathcal{A}}$,
we add the identity matrix $\bm{I}$ to $\bm{\mathcal{A}}_{k,i}$, 
which allows the number of actual message passing steps in the searched meta graph to be flexible.
Moreover, we add the empty matrix $\bm{O}$ to $\bm{\mathcal{A}}_{k, i}$ ($i < k - 1$),
so that if the empty matrix receives the highest weight after search,
we will drop the contribution of $\bm{H}^{(i)}$ to $\bm{H}^{(k)}$ ($i < k - 1$).
This mechanism allows the number of incoming links for $\bm{H}^{(k)}$ in the searched meta graph 
to be flexible in a range of $[1, k]$, and if all $\bm{H}^{(k)}$'s have only one incoming link,
then the searched meta graph becomes a meta path.
These two additional matrices enable us to search for a flexible meta graph in the predefined search space.

Besides,
we notice that in a HIN only node representations of particular types are
related to the downstream task. For example,
when predicting research areas for authors in Figure~\ref{fig:schema},
only author representations are related to the evaluation.
We denote the collection of those edge types
whose target node type is related to the evaluation as $\overline{\bm{\mathcal{A}}}$,
and then candidate edge types for message passing steps that contribute to
final representations $\bm{H}^{(K)}$ should be in $\overline{\bm{\mathcal{A}}}$,
otherwise message passing will have no effect on the nodes related to the evaluation.
To summarize, we define $\bm{\mathcal{A}}_{k,i}$ as:
\begin{equation}
\bm{\mathcal{A}}_{k,i} = 
\begin{cases}
    \bm{\mathcal{A}} \cup \{\bm{I}\}                             & k < K ~\text{and}~ i = k - 1 \\
    \bm{\mathcal{A}} \cup \{\bm{I}\} \cup \{\bm{O}\}             & k < K ~\text{and}~ i < k - 1 \\
    \overline{\bm{\mathcal{A}}}                                  & k = K ~\text{and}~ i = K - 1 \\
    \overline{\bm{\mathcal{A}}} \cup \{\bm{I}\} \cup \{\bm{O}\}  & k = K ~\text{and}~ i < K - 1
\end{cases},
\end{equation}

Even though the task-dependent type constraint 
filters out certain candidate edge types,
the search space is still very large.
For Douban movie recommendation dataset with $11$ edge types (see Appendix B),
when $K = 4$,
to generate user representations
the total number of possible meta graphs is
$(11 + 1)^3 \times (11 + 2)^3 \times 3 \times (3 + 2)^3 \approx 1.4 \times 10^9$,
which makes it necessary to design an efficient search algorithm.

\begin{remark}
To integrate Definition~\ref{def:meta_graph} into the message passing framework,
we omit the constraint of a single source node type, and generalize a meta graph
to a more flexible subgraph defined on the network schema which only
requires a single target node type.
\end{remark}

\subsection{Search Objective}
\label{sec:obj}
At the search stage,
to differentiate the importance of candidate edge types 
for each link in the DAG,
we introduce the architecture parameters $\lambda_{k,i}^m$ 
($m=1,\dots,|\bm{\mathcal{A}}_{k,i}|$) to mix them in $\bar{f}_{k,i}(\cdot)$:
\begin{align}
\alpha_{k, i}^m &= 
\exp(\lambda_{k, i}^m)
	/ \sum\nolimits_{m'=1}^{|\bm{\mathcal{A}}_{k,i}|} \exp(\lambda_{k, i}^{m'}), 
\label{eq:normalize} 
\\
\Bar{f}_{k, i}(\bm{H}^{(i)}; 
& \bm{\mathcal{A}}_{k, i}) 
= 
\sum\nolimits_{m=1}^{|\bm{\mathcal{A}}_{k, i}|}
\alpha_{k,i}^m \cdot f(\bm{H}^{(i)};\bm{\mathcal{A}}_{k,i}^m) 
\label{eq:mixed_msg},
\end{align}
where $f(\bm{H}^{(i)}; \bm{\mathcal{A}}_{k,i}^m)$ denotes to aggregate information $\bm{H}^{(i)}$ from neighbors defined
by $\bm{\mathcal{A}}_{k,i}^m$, the $m$-th edge type in $\bm{\mathcal{A}}_{k,i}$. 

Let $\bm{\omega}$ denote the conventional parameters (hidden weight matrices and biases) of our model 
and $\bm{\lambda}$ be the collection of all architecture parameters.
Following the NAS framework~\cite{liu2018darts,yao2019differentiable}, 
at the search stage,
we aim to solve a bi-level optimization problem:
\begin{equation}
\min\nolimits_{\bm{\lambda}} \mathcal{L}_{\text{val}}(\bm{\omega}^*(\bm{\lambda}), \bm{\lambda}), 
\; \text{s.t.} \; 
\bm{\omega}^*(\bm{\lambda})
= \arg\min\nolimits_{\bm{\omega}} \mathcal{L}_{\text{tra}}(\bm{\omega}, \bm{\lambda}),
\label{eq:bilevel}
\end{equation}
where $\mathcal{L}_{\text{val}}$ and $\mathcal{L}_{\text{tra}}$ represent the validation loss and the training loss, respectively.
Next,
we show two popular tasks over HINs which will be examined in experiments.

\subsubsection{Node Classification}
Node representations output
by the heterogeneous message passing layer can be 
adapted for different downstream tasks.
For the node classification task,
we append a linear layer after the heterogeneous message passing layer,
which reduces the hidden dimension 
to the number of classes,
i.e.,
\begin{align*}
\bar{\bm{Y}} 
& = \text{softmax}(\bm{Z}\bm{W}_o), 
\end{align*}
where $\bm{W}_o \in \mathbb{R}^{d \times C}$ is the output weight matrix 
and $C$ is the number of classes.
Then, 
we use cross-entropy loss
over all labeled nodes as
\begin{align*}
	\mathcal{L} 
	& = - \sum\nolimits_{v \in \mathcal{V}_L}\sum\nolimits_{c=1}^C \bm{y}_v[c] \log \bar{\bm{y}}_v[c],
\end{align*}
where 
$\mathcal{V}_L$ denotes the set of labeled nodes, 
$\bm{y}_v$ is a one-hot 
vector indicating the label of node $v$,
and  $\bar{\bm{y}}_v$ is the predicted label
for the corresponding node in $\bar{\bm{Y}}$.


\subsubsection{Recommendation}
For the recommendation task,
we use the following loss:
\begin{equation}
	\mathcal{L} 
	\! = \! - \! 
	\sum\nolimits_{(u, v) \in \Omega^+}
	\log \sigma(\textbf{z}_u^\top \textbf{z}_v) -
	\sum\nolimits_{(u', v') \in \Omega^-}
	\log \sigma(- ~\textbf{z}_{u'}^\top \textbf{z}_{v'}),
\end{equation}
where $\Omega^+$ and $\Omega^-$ denote the set of observed positive pairs and the set of negative pairs respectively, and $\sigma$ denotes the sigmoid function.
$\textbf{z}_u, \textbf{z}_v, \textbf{z}_{u'}, \textbf{z}_{v'}$ are node representations output by the heterogeneous message passing layer.

\subsection{Search Algorithm}
\label{sec:alg}

The architecture parameters enable us
to mix candidate meta graphs as in~\eqref{eq:mixed_msg}.
However,
naive training with mixed operations,
as existing one-shot NAS algorithms (e.g., DARTS~\cite{liu2018darts}),
has two potential issues.
First, to evaluate training and validation losses
we need to compute message passing once for each 
candidate edge type in the mixed operation~\eqref{eq:mixed_msg}
and for each link in the DAG search space,
which is not efficient enough.
Second,
training with mixed operations is not well consistent with
our final goal,
i.e.,
to derive a single promising meta graph,
because different candidate message passing steps in~\eqref{eq:mixed_msg} 
may correlate with each other~\cite{bender2018understanding} and 
thus metrics evaluated using the mixed model
at the search stage cannot well indicate the performance of
the derived model.
Such issues motivate us to propose a new search algorithm to facilitate the search
process.

\subsubsection{First-order Approximation}

While $\bm{\omega}^*(\bm{\lambda})$ is coupled with $\bm{\lambda}$ in \eqref{eq:bilevel}, 
such an issue can be empirically addressed by approximating $\bm{\omega}^*(\bm{\lambda})$ with the $\bm{\omega}$ dynamically maintained during training ~\cite{liu2018darts, yao2019differentiable, xie2018snas}.
Then,
the optimization is performed with alternate updates, where
in the first phase of each iteration $\bm{\lambda}$ is fixed and $\bm{\omega}$ is updated based on $\partial \mathcal{L}_{\text{tra}} / \partial \bm{\omega}$, 
and in the second phase $\bm{\lambda}$ is updated based on $\partial \mathcal{L}_{\text{val}} / \partial \bm{\lambda}$ while $\bm{\omega}$ being fixed.

\subsubsection{Differentiating Argmax Operation}
To align the search procedure with the goal of deriving a good meta graph,
in each iteration we consider transforming $\alpha_{k,i}^m$
in~\eqref{eq:mixed_msg} into $\bar{\alpha}_{k,i}^m$
for loss evaluation:
\begin{equation}
\label{eq:sample}
\bar{\alpha}_{k,i}^m = 
\begin{cases}
 \alpha_{k,i}^m & m = m^* \\
  0 & \text{otherwise}
\end{cases},
\end{equation}
where $m^* = \arg\max_m \alpha_{k,i}^m$.
With \eqref{eq:sample},
only the message passing step associated with the maximum weight in~\eqref{eq:mixed_msg} needs to be computed to obtain  $\partial \mathcal{L}_{\text{tra}} / \partial \bm{\omega}$. 
However, 
when we are updating $\bm{\lambda}$,
the transformation from $\bm{\alpha}_{k,i}$ to $\bar{\bm{\alpha}}_{k,i}$ in~\eqref{eq:sample} is not differentiable and prevents gradients from 
propagating backwards to $\bm{\alpha}_{k,i}$ and $\bm{\lambda}_{k,i}$.
We address this issue with a softmax approximation:
\begin{equation}
\label{eq:softmax}
\bar{\alpha}_{k,i}^m = 
\lim\nolimits_{t\rightarrow0^+} \alpha_{k,i}^m 
\cdot 
h(m;t),
\end{equation}
where $h(m;t) = 
\exp \left( \nicefrac{\alpha_{k,i}^m}{t} \right) / 
\sum\nolimits_{p = 1}^{|\bm{\mathcal{A}}_{k,i}|} \exp\left( \nicefrac{\alpha_{k,i}^p}{t} \right)$, and $t > 0$ is a temperature parameter.
Then \eqref{eq:softmax} enables us to compute $\partial \mathcal{L}_{\text{val}} / \partial \bm{\alpha}_{k,i}$ approximately, 
and as shown by Proposition~\ref{pr:grad}, 
when $t \rightarrow 0^+$, 
only the gradient 
with respect to the maximum weight 
$\alpha_{k,i}^{m^*}$ in~\eqref{eq:mixed_msg} is non-zero
and can further propagate backwards to $\bm{\lambda}_{k,i}$ via~\eqref{eq:normalize}.
We provide the proof of Proposition~\ref{pr:grad} in Appendix A.

\begin{proposition}\label{pr:grad}
Let $m^* = \arg\max_m \alpha_{k,i}^m$. 
Then, 
we have,
i)
$\lim_{t \rightarrow 0^+}$ 
$\partial \mathcal{L}_{\text{val}} / \partial \alpha_{k,i}^m = 0$ 
for $m \neq m^*$;
ii)
$\lim_{t \rightarrow 0^+}$
$\partial \mathcal{L}_{\text{val}} / \partial \alpha_{k,i}^{m^*}
= \partial \mathcal{L}_{\text{val}} / \partial\bar{\alpha}_{k,i}^{m^*}$.
\end{proposition}

\subsubsection{Complete Algorithm}
The search algorithm for \model \, based on the above discussion is described in Algorithm~\ref{alg:train}.
Note that the softmax approximation in~\eqref{eq:softmax} is only employed for intermediate analysis, 
and we do not need it in the final algorithm by driving $t$ to $0^+$.
In the $i$-th iteration at the search stage, 
we sample one candidate message passing step,
indicated by $\alpha_{k,i}^*$,
for each mixed $\bar{f}_{k,i}(\cdot)$:
\begin{equation}
\label{eq:single_path}
    \alpha_{k,i}^* = 
\begin{cases}
    \alpha_{k,i}^{m^*} & \text{with probability} ~ 1 - \epsilon_i \vspace{0.1cm} \\
    \texttt{rand}(\{\alpha_{k,i}^m\}_{m=1}^{|\bm{\mathcal{A}}_{k,i}|}) & \text{with probability} ~ \epsilon_i
\end{cases},
\end{equation}
where $\texttt{rand}(\cdot)$ means sampling one element from the given set randomly and uniformly.
$\epsilon_i \in (0, 1)$ is a small parameter that encourages exploring different message passing options at the beginning and decreases to $0$ as $i$ increases (using a decay factor of $0.9$ in our experiments).
Forward and backward propagations in the neural network for updating both $\bm{\omega}$ and $\bm{\lambda}$ only involve one sampled message passing 
step for each link in the DAG,
which accelerates the training. 
After search,
to derive the meta graph for evaluation,
we pick the edge type $\bm{\mathcal{A}}_{k,i}^{m^*}$,
which is associated with the maximum weight $\alpha_{k,i}^{m^*}$ in \eqref{eq:mixed_msg},
for each $\bar{f}_{k,i}(\cdot)$ in the DAG search space.
Then,
the message passing step determined by the derived meta graph is
$\bar{f}_{k,i}(\bm{H}^{(i)} ; \bm{\mathcal{A}}_{k,i}) = f(\bm{H}^{(i)} ; \bm{\mathcal{A}}_{k,i}^{m^*})$.

\begin{algorithm}[ht]
\caption{Search algorithm for~\model}
\label{alg:train}
\begin{algorithmic}[1]
\REQUIRE step sizes $\eta_{\bm{\omega}}$, $\eta_{\bm{\lambda}}$, and edge types $\bm{\mathcal{A}}$.
\STATE Initialize parameters $\bm{\omega}$ and 
architecture parameters $\bm{\lambda}$;
\FOR{the $i$-th iteration}
\STATE Compute weights $\bm{\alpha}_{k,i}$ as in~\eqref{eq:normalize} for each $\bar{f}_{k,i}(\cdot)$;
\STATE Sample one candidate message passing step for each $\bar{f}_{k,i}(\cdot)$ as in~\eqref{eq:single_path}. 
The collection of sampled $\alpha_{k,i}^*$'s is denoted as $\bm{\alpha}^*$;
\STATE $\bm{\omega} \leftarrow \bm{\omega} - \eta_{\bm{\omega}} 
\frac{\partial \mathcal{L}_{\text{tra}}(\bm{\omega}, \bm{\alpha}^*)}{\partial \bm{\omega}}$;
\STATE $\bm{\lambda} \leftarrow \bm{\lambda} - \eta_{\bm{\lambda}}
\frac{\partial \mathcal{L}_{\text{val}}(\bm{\omega}, \bm{\alpha}^*)}{\partial \bm{\alpha}^*}
\frac{\mathrm{d}\bm{\alpha}^*}{\mathrm{d}\bm{\lambda}}$ (by Proposition~\ref{pr:grad} and~\eqref{eq:normalize});
\ENDFOR
\STATE Derive a meta graph by selecting $\bm{\mathcal{A}}_{k,i}^{m^*}$ for each $\bar{f}_{k,i}(\cdot)$;
\RETURN the derived meta graph.
\end{algorithmic}
\end{algorithm}

\subsubsection{Comparison with Other One-shot NAS Algorithms}
\label{sec:alg:comp}

Table~\ref{tab:algs} summarizes the difference between our search algorithm 
and some popular one-shot NAS algorithms~\cite{liu2018darts, xie2018snas, yao2019differentiable}. 
In Algorithm~\ref{alg:train},
evaluation of derivatives with respect to both parameters $\bm{\omega}$
and architecture parameters $\bm{\lambda}$
only involves one candidate message passing step for each mixed operation,
and thus speeds up the search process.

\begin{table}[ht]
	\centering
	\caption{Comparison of our search algorithm with existing one-shot NAS algorithms. 
		\ding{51} and \ding{55} indicate whether update needs to compute all candidates of 
		a mixed operation.}
	\vspace{-10px}
	\label{tab:algs}
	\begin{tabular}{c c c}
		\toprule
		Method       &  Updating $\bm{\omega}$   & Updating $\bm{\lambda}$   \\ \midrule
		DARTS~\cite{liu2018darts}        & \ding{51}            & \ding{51}           \\
		SNAS~\cite{xie2018snas}         & \ding{51}            & \ding{51}           \\
		NASP~\cite{yao2019differentiable}         & \ding{55}            & \ding{51}           \\
		Ours         & \ding{55}            & \ding{55}           \\ \bottomrule
	\end{tabular}
\end{table}

The most related one-shot NAS method to ours is NASP~\cite{yao2019differentiable}
which also transforms $\bm{\alpha}_{k,i}$ into one-hot $\bar{\bm{\alpha}}_{k,i}$
as in~\eqref{eq:sample}.
However, 
NASP performs proximal gradient descent which needs
to evaluate $\partial \mathcal{L}_{\text{val}} / \partial \bar{\bm{\alpha}}_{k,i}$.
Note that $\partial \mathcal{L}_{\text{val}} / \partial \bar{\alpha}^m_{k,i}$ ($m \neq m^*$) is
not zero even though the value of $\bar{\alpha}_{k,i}^m$ ($m \neq m^*$) itself is zero.
Therefore,
NASP still needs to compute the candidate operation 
associated with $\bar{\alpha}_{k,i}^m$ ($m \neq m^*$) to obtain
$\partial \mathcal{L}_{\text{val}} / \partial \bar{\alpha}^m_{k,i}$ ($m \neq m^*$).
Different from NASP,
we differentiate the transformation of~\eqref{eq:sample} through a softmax approximation,
in order to compute $\partial \mathcal{L}_{\text{val}} / \partial \bm{\alpha}_{k,i}$ directly.
The Gumbel-softmax trick~\cite{xie2018snas} also uses softmax with a temperature parameter $t$
to approximate $\arg\max$, 
however,
in this paper we point out by Proposition~\ref{pr:grad}
that when $t \rightarrow 0^+$ only the gradient with respect
to $\alpha_{k,i}^{m^*}$ is non-zero, and thus avoid 
computing the operation associated with $\alpha_{k,i}^m$ ($m \neq m^*$).
Besides, we do not need to design an annealing schedule for $t$ like~\cite{xie2018snas}.

\section{Experiments}
\label{sec:exp}

In this section,
we demonstrate the effectiveness
and efficiency of \model~
through extensive experiments on
different kinds of heterogeneous datasets.
We want to address the following questions:
\begin{itemize}[leftmargin=*]
    \item How does \model~perform in the node classification task
    and the recommendation task compared with state-of-the-art baselines?
    \item How efficient is our proposed search algorithm
    compared against 
    genetic search and popular one-shot NAS algorithms?
    \item How efficient is \model~compared against methods which can learn meta paths implicitly?
    \item What architectures are discovered by \model~over different real-world heterogeneous datasets?
\end{itemize}

\subsection{Experimental Setup}

\subsubsection{Datasets}

We evaluate \model~on
two popular tasks~\cite{yang2020heterogeneous}: node classification and recommendation.
In the node classification task,
we want to predict labels for nodes of a particular type,
based on the network structure and node features.
We use three real-world datasets: 
DBLP,
ACM and IMDB.
DBLP contains three types of nodes: papers (\textsf{P}),
authors (\textsf{A}) and conferences (\textsf{C}), and authors are labeled
by their research areas.
ACM contains three types of nodes: papers (\textsf{P}),
authors (\textsf{A}) and subjects (\textsf{S}), and papers are labeled by research 
areas.
IMDB contains three types of nodes: movies (\textsf{M}),
actors (\textsf{A}) and directors (\textsf{D}), and labels are genres of movies.
Nodes in these datasets have bag-of-words representations as input features.
We follow the splits provided by
GTN~\cite{NIPS2019_9367}\footnote{\url{https://github.com/seongjunyun/Graph_Transformer_Networks}}.
Statistics of these datasets are summarized in Appendix~B.

In the recommendation task,
we want to predict links between source nodes (e.g., users)
and target nodes (e.g., items).
We consider three commonly used heterogeneous recommendation datasets\footnote{\url{https://github.com/librahu/HIN-Datasets-for-Recommendation-and-Network-Embedding}}:
Yelp, Douban movie (denoted as ``Douban'' in the sequel) and Amazon.
Yelp is a platform where users review businesses.
Douban is a social media community where users share reviews about movies.
Amazon is a large e-commerce platform which contains
users' ratings for items.
All the three datasets contain rich semantic relations, and their statistics 
are summarized in Appendix~B.
We convert ratings into binary class labels.
Specifically,
we randomly pick $\%50$ of ratings which are 
higher than $3$ as positive pairs,
i.e.,
with label ``1'',
and all ratings which are lower than $4$ serve as negative pairs,
i.e.,
with label ``0''. 
Positive pairs are then randomly split into
a training set, a validation set and a test set
according to a ratio of $3$:$1$:$1$.
We also randomly split negative pairs 
so that in each set
the number of positive pairs and the number of negative pairs 
are the same.
If the total number of negative pairs is less than
the total number of positive pairs,
we treat a source node and a target node which are unconnected
in the original network as an additional negative pair,
in order to balance two classes.
To avoid the label leakage issue~\cite{ijcai2019-0592},
both positive pairs and negative pairs are disconnected in the original network.
Nodes in these three datasets are not associated with
attributes,
so we use one-hot IDs as input features.

\begin{table*}[ht]
	\centering
	\caption{Macro F1 scores (\%) on the node classification task.}
	\label{tab:nc}
	\vspace{-10px}
	\begin{tabular}{ccccccccc}
		\toprule
		     &    metapath2vec       &  GCN               &  GAT            &  HAN               & MAGNN              &  GTN              & HGT               & \textbf{\model}        \\ \midrule
		DBLP &    89.93$\pm$0.45     & 90.46$\pm$0.41   &  93.92$\pm$0.28   & 92.13$\pm$0.26     & 92.81$\pm$0.30     & 93.98$\pm$0.32    & 93.67$\pm$0.22    & \textbf{94.45$\pm$0.15}  \\
		ACM  &    67.13$\pm$0.50     & 92.56$\pm$0.20   &  92.50$\pm$0.23   & 91.20$\pm$0.25     & 91.15$\pm$0.19     & 92.62$\pm$0.17    & 91.83$\pm$0.23    & \textbf{92.65$\pm$0.15}  \\
		IMDB &    40.82$\pm$1.48     & 55.19$\pm$0.99   &  53.37$\pm$1.27   & 55.09$\pm$0.67     & 56.44$\pm$0.63     & 59.68$\pm$0.72    & 59.35$\pm$0.79    & \textbf{61.04$\pm$0.56}  \\ \bottomrule
	\end{tabular}
\end{table*}

\begin{table*}[ht]
	\centering
	\caption{AUC (\%) on the recommendation task.}
	\label{tab:lp}
	\vspace{-10px}
	\setlength\tabcolsep{3pt}
	\begin{tabular}{cccccccccc}
		\toprule
		              &    metapath2vec          &  GCN             &  GAT              &  HAN           & MAGNN          & GEMS               &  GTN               &  HGT               &\textbf{\model} \\ \midrule
		Amazon        &    58.17$\pm$0.14        & 66.64$\pm$1.00   &  55.70$\pm$1.13   & 67.35$\pm$0.11 & 68.26$\pm$0.09 & 70.66$\pm$0.14     &  71.82$\pm$0.18    &  74.75$\pm$0.08    &\textbf{75.28$\pm$0.08}  \\
		Yelp          &    51.98$\pm$0.14        & 58.98$\pm$0.52   &  56.55$\pm$0.05   & 64.28$\pm$0.20 & 64.73$\pm$0.24 & 65.12$\pm$0.27     &  66.27$\pm$0.31    &  68.07$\pm$0.35    &\textbf{68.77$\pm$0.13}  \\
		Douban        &    51.60$\pm$0.07        & 77.95$\pm$0.05   &  77.58$\pm$0.33   & 82.65$\pm$0.08 & 82.44$\pm$0.17 & 83.00$\pm$0.05     &  83.26$\pm$0.10    &  83.38$\pm$0.06    &\textbf{83.78$\pm$0.09}  \\ \bottomrule
	\end{tabular}
\end{table*}

\subsubsection{Baselines}
We compare against a random walk based network embedding method
metapath2vec~\cite{dong2017metapath2vec}\footnote{\url{https://ericdongyx.github.io/metapath2vec/m2v.html}},
two homogeneous GNNs: GCN~\cite{kipf2016semi} and GAT~\cite{velivckovic2017graph}\footnote{
	We use implementations from PyTorch Geometric (PyG)~\cite{Fey/Lenssen/2019} for GCN and GAT.},
two heterogeneous GNNs which require manually designed meta paths:
HAN~\cite{wang2019heterogeneous}
\footnote{We implement with PyG following the authors' TensorFlow code at \url{https://github.com/Jhy1993/HAN}.}
and MAGNN~\cite{fu2020magnn}\footnote{\url{https://github.com/cynricfu/MAGNN}},
and two heterogeneous GNNs which 
use attention to implicitly learn meta paths:
GTN~\cite{NIPS2019_9367}\footnote{\url{https://github.com/seongjunyun/Graph_Transformer_Networks}} 
and HGT~\cite{hu2020heterogeneous}\footnote{\url{https://github.com/acbull/pyHGT}}.
In the recommendation task,
we also compare against GEMS~\cite{han2020genetic}\footnote{\url{https://github.com/0oshowero0/GEMS}} 
which utilizes the
evolutionary algorithm to search for meta graphs
between source node type and target node type.
%
All heterogeneous GNN baselines and our own model are also summarized in Table~\ref{tab:hingnn}.
Except metapath2vec,
all methods are implemented using PyTorch~\cite{NEURIPS2019_bdbca288}.

\subsubsection{Evaluation Metrics}
In the node classification task,
we use macro F1 score as the evaluation metric;
in the recommendation task,
we use AUC (area under the ROC curve) as the evaluation metric.

\subsubsection{Hyper-parameters}
Due to randomness of initialization and sampling as in~\eqref{eq:single_path},
for each dataset we run the search algorithm three times with
different random seeds and 
then derive the meta graph for final evaluation from the run which achieves
the best validation performance.
In the node classification task,
we run the search algorithm for $50$ epochs,
while in the recommendation task,
we run it for $100$ epochs.
At the search stage,
we train $\bm{\lambda}$ using Adam with a learning rate of $3e-4$,
and we train $\bm{\omega}$ using Adam with a learning rate of $0.005$
and weight decay of $0.001$.
We set $K = 4$ for all datasets,
which is the same as the length of meta paths 
learned by GTN~\cite{NIPS2019_9367}.
Since meta paths and meta graphs, either manually designed~\cite{wang2019heterogeneous, fu2020magnn} or learned~\cite{NIPS2019_9367, han2020genetic},
enlarge GNN's receptive field,
we use $4$ layers for GCN, GAT and HGT
to keep a fair comparison.
The hidden dimension is set to $64$ for all baselines and our model.
The number of attention heads is $8$ for GAT, HAN, MAGNN and HGT,
so the hidden dimension for each head is $8$.
The dimension of semantic-level attention vector for HAN~\cite{wang2019heterogeneous}
and MAGNN~\cite{fu2020magnn} is $128$.
All GNN methods are trained using full batch,
for $100$ epochs in the node classification task with early stopping,
and for $200$ epochs in the recommendation task.
The optimizer is Adam and other hyper-parameters (learning rate, weight decay and input dropout) are tuned according to the validation performance.
For metapath2vec,
we follow its default hyper-parameter setting.
For GEMS,
we run the evolutionary algorithm for $100$ generations with a population size of $20$.
Experiments are conducted on a single RTX $2080$ Ti GPU with $11$GB memory, except that we parallelize GEMS on more than one GPU to speed up the genetic search.


\subsection{Node Classification}
Table~\ref{tab:nc} shows the macro F1 scores of different methods
on the node classification task.
For each method,
we report the average score and standard deviation of $10$ runs with different random seeds.
First,
GAT yields highly competitive performance on DBLP and ACM compared with heterogeneous GNN baselines.
We conjecture this is because GAT can exploit semantic information by
learning to assign different importance to different node types.
Second,
HAN and MAGNN which rely on hand-designed meta paths do not obtain desirable performance compared 
with GTN, HGT and \model,
and they can even perform worse than homogeneous GNNs.
This indicates that hand-designed rules are limited in mining task-dependent semantic information and
may even cause adverse effects.
Finally,
\model~consistently achieves the best performance on all three datasets,
which demonstrates the importance of automatically utilizing task-dependent semantic information in HINs.
Specifically,
the performance improvement over GTN further validates that meta graphs 
are more powerful to capture semantic information than meta paths.

%

\subsection{Recommendation}
\label{sec:lp}
In Table~\ref{tab:lp},
we report the AUC of \model~
and baselines.
The results are the average of $10$ runs with different random seeds.
The HINs used for
the recommendation task are larger and have richer semantic relations than 
those used for the node classification task,
and we observe that the performance of homogeneous GNNs
is poor compared with heterogeneous GNNs which are tailored to
utilize semantic information.
HAN and MAGNN show performance improvement over homogeneous GNNs
by leveraging 
meta paths.
GEMS outperforms HAN and MAGNN by searching for a meta graph 
to capture semantic proximity between source nodes and target nodes,
but it is too inefficient to fully explore the search space.
Moreover,
GEMS ignores representations of intermediate nodes along the meta graph,
which limits its capacity.
Again,
our method is the best on all three datasets
and exhibits significant improvement over the most competitive baseline HGT,
which demonstrates that a task-dependent meta graph is superior to a much more
complex architecture in representation learning on HINs.

\subsection{Efficiency Analysis}

\subsubsection{Search Stage}\label{sec:seaeff}
In Table~\ref{tab:cost}, 
we compare the search cost of our search algorithm (Algorithm~\ref{alg:train}),
measured in GPU minutes,
against genetic search and other popular one-shot NAS algorithms (Table~\ref{tab:algs}),
on the recommendation datasets.
For one-shot search algorithms,
we report the total search cost of three runs.
We also report the evaluation cost which requires training the searched model from scratch once.
The genetic search algorithm evaluates each candidate model individually,
so it is extremely time consuming.
On the contrary,
one-shot search algorithms combine candidate models in a super model
and reduce the search cost by dozens of times.
Compared with existing one-shot NAS algorithms~\cite{liu2018darts, yao2019differentiable, xie2018snas},
our search algorithm significantly improves the search efficiency,
so that the total search cost is on a par with the evaluation cost.
The efficient search algorithm enables our framework \model~to be applied to large scale heterogeneous datasets.

\begin{table}[ht]
	\centering
	\caption{Search cost compared against evaluation cost, measured in GPU minutes.
	For one-shot NAS methods, 
	we also report in parentheses the time required to update $\bm{\lambda}$.}
	\vspace{-10px}
	\label{tab:cost}
	\begin{tabular}{ccccc}
		\toprule
		~                               & ~                                    & Amazon              & Yelp                      & Douban      \\ \midrule
		Genetic                         & GEMS~\cite{han2020genetic}           & 800                 & 1500                      & 2000              \\ \midrule
		\multirow{4}{*}{One-shot}       & DARTS~\cite{liu2018darts}            & 8.9 (6.9)           & 15.8 (12.2)               & 19.4 (15.0)                 \\
		~                               & SNAS~\cite{xie2018snas}              & 3.9 (1.9)           & 7.0  (3.4)                & 8.6 (4.2)          \\
		~                               & NASP~\cite{yao2019differentiable}    & 2.3 (1.9)           & 4.0  (3.4)               & 5.0 (4.2)        \\
		~                               & Ours                                 & \textbf{0.7 (0.3)}  & \textbf{1.1 (0.5)}  & \textbf{1.6 (0.8)} \\ \midrule
		\multicolumn{2}{c}{Evaluation}            & 0.5           & 0.7           & 1.0          \\ \bottomrule
	\end{tabular}
\end{table}

\subsubsection{Evaluation Stage}
We study the efficiency of \model~at the evaluation stage,
where the derived model is retrained from scratch,
on two large heterogeneous datasets: Yelp and Douban.
We compare against HGT~\cite{hu2020heterogeneous} which can implicitly learn meta paths by
attention and outperforms
the other baselines in Table~\ref{tab:lp}.
Another strong baseline GTN~\cite{NIPS2019_9367} cannot fit into a single GPU on these 
datasets and requires much more time to train on CPU (nearly $20$ minutes per epoch on Yelp),
so we do not compare against it here.
We also compare against the simplest architecture 
GCN to show the trade off between performance and efficiency.

\begin{figure}[ht]
	\centering
	\subfigure[Yelp.]
	{\includegraphics[width=0.48\columnwidth]{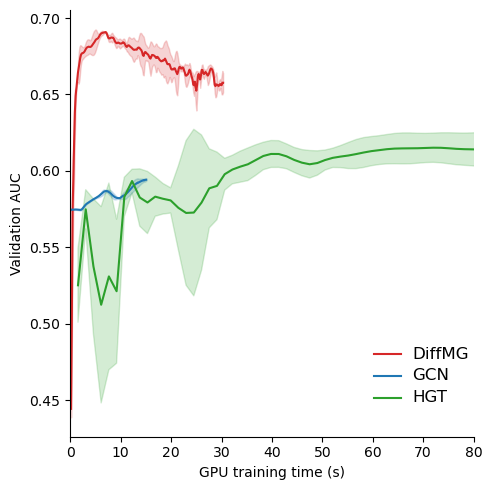}}
	\subfigure[Douban.]
	{\includegraphics[width=0.48\columnwidth]{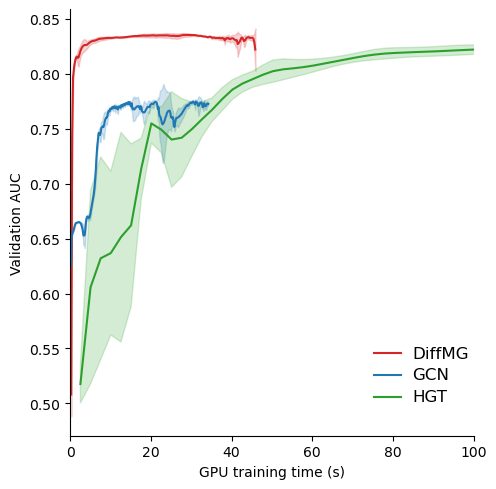}}
	\vspace{-10px}
	\caption{Efficiency of \model~at the evaluation stage. We do not compare against GTN because of OOM on a single GPU.}
	\label{fig:curve}
\end{figure}

In Figure~\ref{fig:curve},
we plot the validation AUC with respect to GPU training time
(measured in seconds) for these methods.
They are trained for $200$ epochs three times.
We observe that \model~reaches its highest validation performance
much faster than HGT.
Compared with GCN,
\model~gains large performance
improvement 
while taking only a mild amount of additional time
to finish training.
Furthermore,
the search cost of \model~
is on a par with training the derived model once
(see Section~\ref{sec:seaeff}),
which demonstrates that \model~
is well applicable to large scale HINs.

\subsection{Visualization}
\label{sec:metagraph}
\begin{figure}[ht]
	\centering
	\subfigure[DBLP.]
	{\includegraphics[width=0.28\columnwidth]{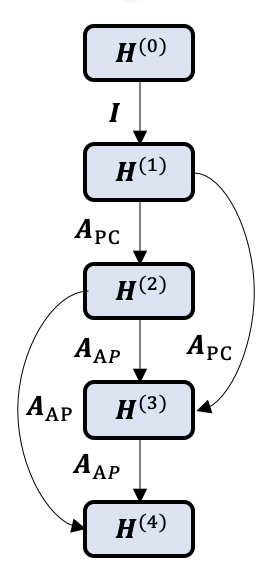}\label{fig:case-dblp}}
	\subfigure[Douban.]
	{\includegraphics[width=0.66\columnwidth]{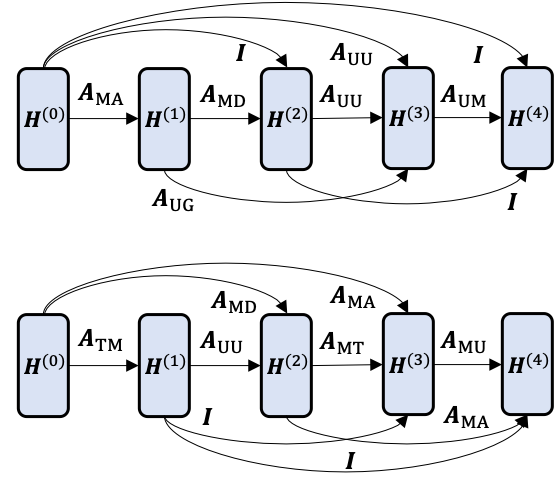}\label{fig:case-douban}}
	\vspace{-10px}
	\caption{Architectures discovered by \model.}
	\label{fig:case}
\end{figure}

We visualize the architectures discovered by \model~on DBLP and Douban in Figure~\ref{fig:case}.
In Figure~\ref{fig:case-dblp},
we observe that both conference nodes (\textsf{C}) and paper nodes (\textsf{P})
are involved in generating representations for authors (\textsf{A}),
which is similar to hand-designed meta paths for academic networks~\cite{dong2017metapath2vec}.
Moreover,
we observe that $\bm{H}^{(3)}$ and $\bm{H}^{(4)}$ have more than one incoming link,
which combines multiple propagation paths together.
This cannot be done by GTN~\cite{NIPS2019_9367} which only learns meta paths via matrix multiplication.
In Figure~\ref{fig:case-douban},
we show both the meta graph (above) whose target node type is user (\textsf{U}) and the meta graph (below)
whose target node type is movie (\textsf{M}).
\model~is able to obtain different meta graphs to utilize semantic information
for different target node types.
In our framework,
each edge type in the derived meta graph corresponds to
message passing along edges of this type in the underlying HIN,
which enables \model~to encode information of intermediate nodes along a meta graph.
On the contrary,
GEMS~\cite{han2020genetic} only propagates information between source nodes and target nodes,
and ignores intermediate nodes along the meta graph.

\subsection{Ablation Study}
In Table~\ref{tab:ablation},
we study the influence of $\epsilon_0$
on the final performance.
Search and evaluation procedures for Table~\ref{tab:ablation}
are the same as Section~\ref{sec:lp}.
We observe that large $\epsilon_0$
does not improve the performance except 
that $\epsilon_0 = 0.5$ works better 
on Amazon than $\epsilon_0 = 0$.
We also show the performance of single-level
optimization which optimizes both $\bm{\omega}$
and $\bm{\lambda}$ with the union of training set
and validation set,
and it matches bi-level optimization.
To validate the second motivation
of our proposed search algorithm,
we study a variant which
optimizes all architecture parameters of a mixed operation 
simultaneously, like DARTS~\cite{liu2018darts}.
Its performance is worse than our search algorithm.

\begin{table}[ht]
    \centering
    \caption{Ablation study results (\%).}
    \label{tab:ablation}
    \vspace{-10px}
    \begin{tabular}{lcccc}
    \toprule
    ~                    &  Amazon                                 & Yelp                       & Douban                \\ \midrule
    \model~($\epsilon_0 = 0$)     &  74.80$\pm$0.09                         & 68.77$\pm$0.13             & 83.78$\pm$0.09        \\ 
    \model~($\epsilon_0 = 0.3$)   &  74.51$\pm$0.17                         & 68.42$\pm$0.09             & 83.58$\pm$0.08        \\ 
    \model~($\epsilon_0 = 0.5$)   &  75.28$\pm$0.08                         & 67.90$\pm$0.15             & 83.45$\pm$0.09        \\ 
    \model~(single-level)         &  74.80$\pm$0.09                         & 68.94$\pm$0.12             & 83.70$\pm$0.03        \\
    \model~(DARTS)                &  74.54$\pm$0.22                         & 66.72$\pm$0.20             & 83.68$\pm$0.05        \\
    \bottomrule
    \end{tabular}
\end{table}

\section{Conclusion}
In this work,
we explore the direction of addressing representation learning on HINs with automated
machine learning.
We present key contributions on both the design of the search space and the search algorithm.
For future work, 
we will study the convergence of our proposed search algorithm.

\bibliographystyle{ACM-Reference-Format}
\balance
\bibliography{ref}

\clearpage
\appendix

\section{Proof}
We provide the proof of Proposition~\ref{pr:grad} as follows.
According to the chain rule, we have:
\begin{align}
\label{eq:a1}
    \frac{\partial \mathcal{L}_{\text{val}}}{\partial \alpha_{k,i}^m}
    = \frac{\partial \mathcal{L}_{\text{val}}}{\partial \bm{H}^{(k)}} \frac{\partial \bm{H}^{(k)}}{\partial \alpha_{k,i}^m}
    = \frac{\partial \mathcal{L}_{\text{val}}}{\partial \bm{H}^{(k)}} \sum_{q = 1}^{|\bm{\mathcal{A}}_{k,i}|} \frac{\partial \bar{f}_{k,i}}{\partial \bar{\alpha}_{k,i}^q} \frac{\partial \bar{\alpha}_{k,i}^q}{\partial \alpha_{k,i}^m}.
\end{align}
Recall that in~\eqref{eq:softmax} we use a softmax function to approximate the discrete transformation of~\eqref{eq:sample}:
\begin{equation}
\label{eq:a2}
    \bar{\alpha}_{k,i}^m \approx \alpha_{k,i}^m \cdot h(m;t),
\end{equation}
where $h(m;t) = 
\exp \left( \nicefrac{\alpha_{k,i}^m}{t} \right) / 
\sum\nolimits_{p = 1}^{|\bm{\mathcal{A}}_{k,i}|} \exp\left( \nicefrac{\alpha_{k,i}^p}{t} \right)$, and $t > 0$ is a temperature parameter.
Based on~\eqref{eq:a2}, we have:
\begin{align}
\label{eq:a3}
    \frac{\partial \bar{\alpha}_{k,i}^q}{\partial \alpha_{k,i}^m} 
    = \delta_{qm}h(q; t) + \alpha_{k,i}^q\frac{\partial h(q;t)}{\partial \alpha_{k,i}^m},
\end{align}
where if $q$ equals $m$, $\delta_{qm} = 1$, otherwise $\delta_{qm} = 0$.
Combining~\eqref{eq:a1} and~\eqref{eq:a3}, and note that $\lim_{t \rightarrow 0^+} \partial h(q;t) / \partial \alpha_{k,i}^m = 0$, then we get:
\begin{align}
    \lim\nolimits_{t \rightarrow 0^+} \frac{\partial \mathcal{L}_{\text{val}}}{\partial \alpha_{k,i}^m}
    &= \lim\nolimits_{t \rightarrow 0^+} \frac{\partial \mathcal{L}_{\text{val}}}{\partial \bm{H}^{(k)}} \frac{\partial \bar{f}_{k,i}}{\partial \bar{\alpha}_{k,i}^m} h(m; t) \\
    &= \lim\nolimits_{t \rightarrow 0^+} \frac{\partial \mathcal{L}_{\text{val}}}{\partial \bar{\alpha}_{k,i}^m} h(m;t).
\end{align}
Since $\lim\nolimits_{t \rightarrow 0^+} h(m^*;t) = 1$, and $\lim\nolimits_{t \rightarrow 0^+} h(m;t) = 0$ for $m \neq m^*$,
we finish the proof.

\section{Statistics of Datasets}

\begin{table}[H]
    \small
	\centering
	\caption{Statistics of HINs for node classification.}
	\vspace{-10px}
	\small
	\begin{tabular}{lccc}
		\toprule
		& DBLP   & ACM    & IMDB    \\ \midrule
		\#~Nodes        & 18405  & 8994   & 12624   \\
		\#~Edges        & 67946  & 25922  & 37288   \\
		\#~Classes      & 4      & 3      & 3       \\
		\#~Edge types   & 4      & 4      & 4       \\
		\#~Training     & 800    & 600    & 300     \\
		\#~Validation   & 400    & 300    & 300     \\
		\#~Testing      & 2857   & 2125   & 2339    \\ \bottomrule
	\end{tabular}
\end{table}

\begin{table}[H]
	\small
	\centering
	\caption{Statistics of HINs for recommendation.}
	\vspace{-10px}
	\begin{tabular}{ccccc}
		\toprule
		Dataset                   &  Relations (A-B)                & \#~A           & \#~B             & \#~A-B           \\ \midrule
		\multirow{5}{*}{Yelp}     &  \textbf{User-Business (U-B)}   & \textbf{16239} & \textbf{14284}   & \textbf{198397}  \\
		~                         &  User-User (U-U)                & 16239          & 16239            & 158590           \\
		~                         &  User-Compliment (U-Co)         & 16239          & 11               & 76875            \\
		~                         &  Business-City (B-C)            & 14284          & 47               & 14267            \\
		~                         &  Business-Category (B-Ca)       & 14284          & 511              & 40009            \\ \midrule
		\multirow{6}{*}{\shortstack{Douban\\movie}} & \textbf{User-Movie (U-M)}  & \textbf{13367} & \textbf{12677} & \textbf{1068278} \\ 
		~   & User-Group (U-G) & 13367 & 2753  & 570047 \\
		~   & User-User  (U-U) & 13367 & 13367 & 4085  \\
		~   & Movie-Actor (M-A) & 12677 & 6311 & 33587 \\
		~   & Movie-Director (M-D) & 12677 & 2449 & 11276 \\
		~   & Movie-Type  (M-T) & 12677 & 38 & 27668 \\ \midrule
		\multirow{4}{*}{Amazon} & \textbf{User-Item (U-I)} & \textbf{6170} & \textbf{2753} & \textbf{195791} \\
		~ & Item-View (I-V) & 2753 & 3857 & 5694 \\
		~ & Item-Category (I-C) & 2753 & 22 & 5508 \\
		~ & Item-Brand (I-B) & 2753 & 334 & 2753 \\ \bottomrule
	\end{tabular}
\end{table}

\end{document}